\documentclass[11pt,a4paper]{article}
\usepackage{mathptmx}
\usepackage{amssymb}
\usepackage{framed}
\usepackage{xcolor,colortbl}
\usepackage{footnote}
\usepackage{enumitem}
\usepackage{xspace}
\usepackage{dblfloatfix}
\definecolor{mygray}{gray}{0.85}
\usepackage{enumerate}
\usepackage[multiple]{footmisc}
\usepackage{cleveref}
\usepackage{authblk}
\usepackage{multirow}
\usepackage{wrapfig}
\usepackage{pgfplotstable}
\pgfplotsset{compat= 1.13}
\usepackage{adjustbox}
\usepackage{amsmath}

\DeclareMathOperator{\softmax}{softmax}
\usepackage[hyperref]{acl2018}
\usepackage{times}
\usepackage{latexsym}
\usepackage{graphicx}
\usepackage{tikz}
\usepackage{subfig}
\usepackage{url}
\usepackage{booktabs}

\newcommand{\mlp}{\textsc{mlp}\xspace}
\newcommand{\bilstm}{\textsc{bilstm}\xspace}

\newcommand{\pos}{\textsc{pos}\xspace}
\newcommand{\wordvec}{\textsc{word2vec}\xspace}

\newcommand{\lr}{\textsc{lr}\xspace}

\newcommand{\rnn}{\textsc{rnn}\xspace}

\newcommand{\lstm}{\textsc{lstm}\xspace}

\newcommand{\R}{\mathbb{R}}

\newcommand{\bilstmatt}{\textsc{bilstm-att}\xspace}
\newcommand{\xbilstmatt}{\textsc{x-bilstm-att}\xspace}
\newcommand{\hbilstmatt}{\textsc{h-bilstm-att}\xspace}
\newcommand{\auc}{\textsc{auc}\xspace}

\aclfinalcopy
\def\aclpaperid{644}

\title{Obligation and Prohibition Extraction Using Hierarchical RNNs}

\author[1,2]{Ilias Chalkidis}
\author[1]{Ion Androutsopoulos}
\author[2]{Achilleas Michos}

\affil[1]{Department of Informatics, Athens University of Economics and Business, Greece}
\affil[2]{Cognitiv+ Ltd., London, UK}

\date{}

\begin{document}
\maketitle
\begin{abstract}
We consider the task of detecting contractual obligations and prohibitions. We show that a self-attention mechanism improves the performance of a \bilstm classifier, the previous state of the 
art for this task, by allowing it to focus on indicative tokens. We also introduce a hierarchical \bilstm, which converts each sentence to an embedding, and processes the sentence embeddings to classify each sentence. Apart from being faster to train, the hierarchical \bilstm outperforms the flat one, even when the latter considers surrounding sentences, because the hierarchical model has a broader discourse view.
\end{abstract}

\section{Introduction}
Legal text processing \cite{Ashley2017} is a growing research area, comprising tasks such as legal question answering \cite{Kim2017}, contract element extraction \cite{Chalkidis2017}, and legal text generation \cite{Alschnerd2017}. We consider \emph{obligation} and \emph{prohibition extraction} from contracts, i.e., detecting sentences (or clauses) that specify what \emph{should} or should \emph{not} happen (Table~\ref{tab:examples}). This task is important for legal firms and legal departments, especially when they process large numbers of contracts to monitor the compliance of each party. Methods that would automatically identify (e.g., highlight) sentences (or clauses) specifying obligations and prohibitions would allow lawyers and paralegals to inspect contracts more quickly. They would also be a step towards populating databases with information extracted from contracts, along with methods that extract contractors, particular dates (e.g., start and end dates), applicable law, legislation references etc.\ \cite{Chalkidis2017b}. 

Obligation and prohibition extraction is a kind of \emph{deontic} sentence (or clause) classification \cite{ONeill2017}. Different firms may use different or finer deontic classes (e.g., distinguishing between payment and delivery obligations), but obligations and prohibitions are the most common coarse deontic classes. Using similar classes, O'~Neill et al.\ \shortcite{ONeill2017} reported that a bidirectional \lstm (\bilstm) classifier \cite{Graves2013} outperformed several others (including logistic regression, \textsc{svm}, AdaBoost,  Random Forests) in legal sentence classification, possibly because long-term dependencies (e.g., modal verbs or negations interacting with distant dependents) are common and crucial in legal texts, and \lstm{s} can cope with long-term dependencies better than methods relying on fixed-size context windows.

\begin{figure}[tb]
  \centering
    \fbox{\includegraphics[width=0.95\columnwidth]{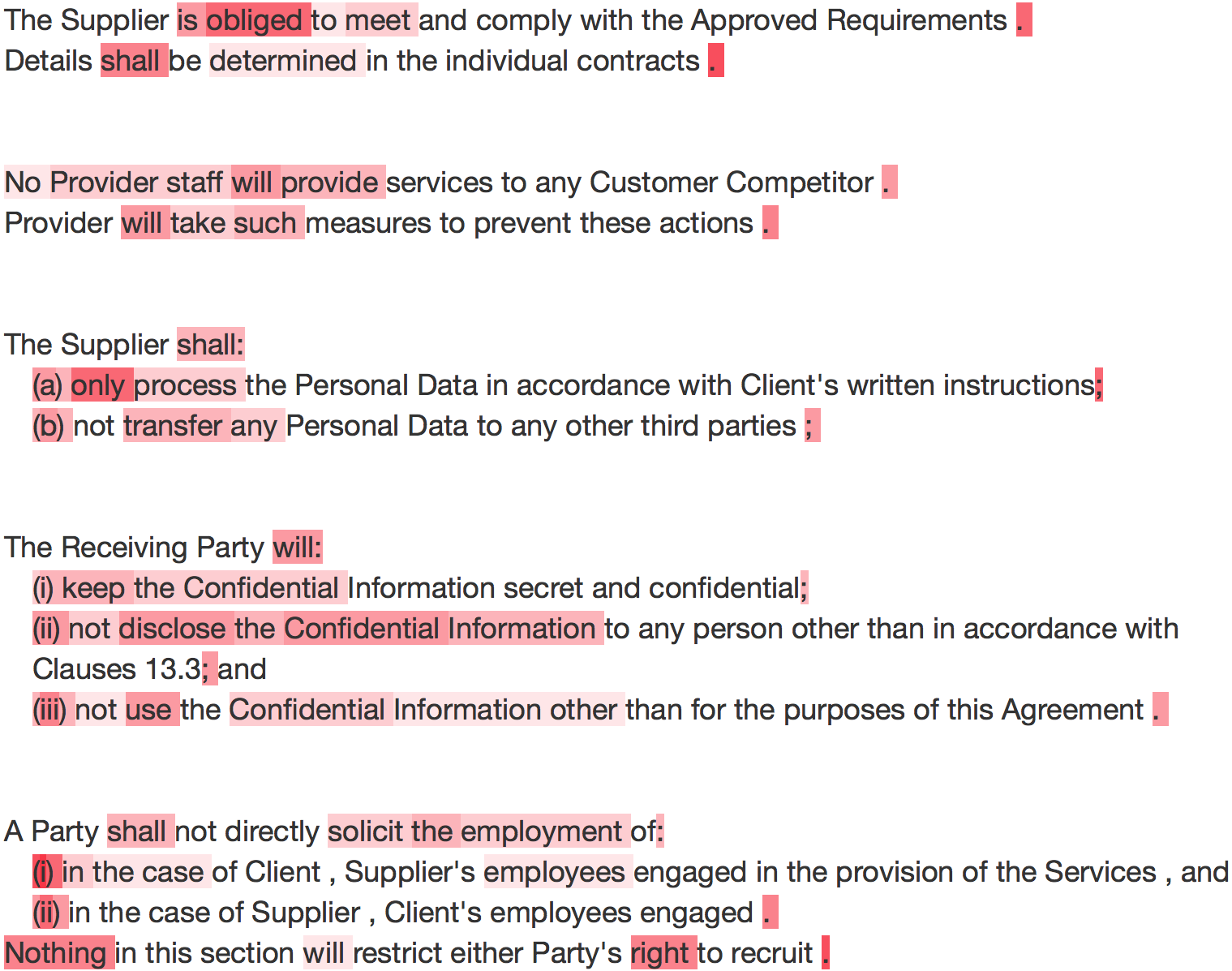}}
  \caption{Heatmap visualizing the attention scores of \bilstmatt for some examples of Table~\ref{tab:examples}.}
  \vspace*{-5mm}
  \label{fig:heatmapFlat}
\end{figure}

\begin{table*}[tb]
\centering
\small
\begin{tabular}{|c|l|p{11cm}|}
\hline
\textbf{No.} &
\textbf{Gold Class} & \textbf{Sentences/Clauses}	\\ \hline
1 & Obligation 	&	The Supplier \textbf{\emph{\underline{is obliged to}}} meet and comply with the Approved Requirements.\\		
& None & Details shall be determined in the individual contracts. \\ \hline
2 & Prohibition				& \textbf{\emph{\underline{No}}} Provider staff \textbf{\emph{\underline{will}}} provide services to any Customer Competitor. \\
  & Obligation  &  Provider \textbf{\emph{\underline{will}}} take such measures to prevent these actions. \\ \hline
3 & Prohibition 			&	 Provider \textbf{\emph{\underline{is not entitled}}} to suspend this Agreement prior to the lapse of the fifth 
year.\\ \hline
4 & Oblig./Prohib.\ List Intro &	The Supplier \textbf{\emph{\underline{shall}}}: \\
  & Obligation List Item	&   \textbf{\emph{\underline{(a)}}} only process the Personal Data in accordance with Client's written instructions; \\ 
 & Prohibition	List Item			&   \textbf{\emph{\underline{(b) not}}} transfer any Personal Data to any other third parties; \\ \hline
5 & Oblig./Prohib.\ List Intro & The Receiving Party \textbf{\emph{\underline{will}}}: \\
  & Obligation List Item	& \textbf{\emph{\underline{(i)}}} keep the Confidential Information secret and confidential;  \\
  & Prohibition	List Item		& \textbf{\emph{\underline{(ii) not}}} disclose the Confidential Information to any person other than in accordance with Clauses 13.3; and \\
  &	Prohibition	List Item			& \textbf{\emph{\underline{(iii) not}}} use the Confidential Information other than for the purposes of this Agreement. \\ \hline 	
6 & Oblig./Prohib.\ List Intro		&  A Party \textbf{\emph{\underline{shall not}}} directly solicit the employment of: \\
  & Prohibition	List Item			& \textbf{\emph{\underline{(i)}}} in the case of Client, Supplier's employees engaged in the provision of the Services, \\
  & Prohibition	List Item			& \textbf{\emph{\underline{(ii)}}} in the case of Supplier, Client's employees engaged. \\
  & None 			&  Nothing in this section will restrict either Party's right to recruit. \\ \hline
\end{tabular}
\caption{Examples of sentences and clauses, with human annotations of classes. Terms that are highly indicative of the classes are shown in bold and underlined here, but are not marked by the annotators.} 
\label{tab:examples}
\end{table*}

We improve upon the work of O' Neill et al.\ \shortcite{ONeill2017} in four ways. First, we show that self-attention \cite{Yang2016} improves the performance of the \bilstm classifier, by allowing the system to focus on indicative words (Fig~\ref{fig:heatmapFlat}). 
Second, we introduce a hierarchical \bilstm, where a first \bilstm processes each sentence word by word producing a sentence embedding, and a second \bilstm processes the sentence embeddings to classify each sentence. The hierarchical \bilstm is similar to Yang et al.'s \shortcite{Yang2016}, but classifies sentences, not entire texts (e.g., news articles or product reviews).
It outperforms a flat \bilstm that classifies each sentence independently, even when the latter considers neighbouring sentences, because the hierarchical \bilstm has a broader view of the discourse. Third, we experiment with a dataset an order of magnitude larger than the dataset of O' Neill et al.
Fourth, we introduce finer classes (Tables~\ref{tab:examples}--\ref{tab:dataset}), which fit better the target task, where nested clauses are frequent.

\section{Data} \label{sec:dataset}

\begin{table}
\centering
\small
\begin{tabular}{|c|c|c|c|}
\hline
\textbf{Gold Class} &  \textbf{Train} & \textbf{Dev} & \textbf{Test} \\ \hline
None					&  15,401	& 3,905		& 4,141  \\ \hline
Obligation 				&  11,005	& 2,860		& 970	\\ \hline
Prohibition 			&  1,172 	& 314		& 108 	\\ \hline
Obligation List Intro 	&  828 		& 203 		& 70 	\\ \hline
Obligation List Item 	&  2888		& 726		& 255 	\\ \hline
Prohibition List Item	&  251		& 28		& 19    \\ \hline
{\bf Total}				&  {\bf 31,545}	& {\bf 8,036}		& {\bf 5,563}	\\ \hline
\end{tabular}
\caption{Sentences/clauses after sentence splitting.}
\label{tab:dataset}
\end{table}

We experimented with a dataset containing 6,385 training, 1,595 development, and 1,420 test sections (articles) from the main bodies (excluding introductions, covers, recitals) of 100 randomly selected English service agreements.\footnote{The splitting of the dataset into training, development, and test subsets was performed by first agglomeratively clustering all sections (articles) based on Levenshtein distance, and then assigning entire clusters to the training, development, or test subset, to avoid having similar sections (e.g., based on boilerplate clauses) in different subsets.}
The sections were preprocessed by a sentence splitter, which in clause lists (Examples 4--6 in Table~\ref{tab:examples}) treats the introductory clause and each nested clause as separate sentences, since each nested clause may belong in a different class.\footnote{We use \textsc{nltk}'s splitter (\url{http://www.nltk.org/}), with additional post-processing based on regular expressions.}

The  splitter produced 31,545 training, 8,036 development, and 5,563 test sentences/clauses.\footnote{There are at most 15 sentences/clauses per section in the training set. We hope to make the dataset, or a similar anonymized one, publicly available in the near future, but the dataset is currently not available due to confidentiality issues.} Table~\ref{tab:dataset} shows their distribution in the six gold (correct) classes. Each section was annotated by a single law student (5 students in total). All the annotations were checked and corrected by a single paralegal expert, who produces annotations of this kind on a daily basis, based on strict guidelines of the firm that provided the data.

We used pre-trained 200-dimensional word embeddings and pre-trained 25-dimensional \pos tag embeddings, obtained by applying \wordvec \cite{MikolovSCCD13} to approx.\ 750k and 50k English contracts, respectively, as in our previous work \cite{Chalkidis2017}. We also pre-trained 5-dimensional token shape embeddings (e.g., all capitals, first letter capital, all digits), obtained as in our previous work \cite{Chalkidis2017b}. Each token is represented by the concatenation of its word, \pos, shape embeddings (Fig.~\ref{fig:encoder},  bottom). Unknown tokens are mapped to pre-trained \pos-specific `unk' embeddings (e.g., `unk-n', `unk-vb'). The dataset of Table~\ref{tab:dataset} has no overlap with the corpus of contracts that was used to pre-train the embeddings.

\section{Methods}

\smallskip
\noindent\textbf{\bilstm}:
The first classifier we considered processes a single sentence (or clause) at a time. It feeds the concatenated word, \pos, shape embeddings ($e_1, \dots, e_n \in \R^{230}$) of the tokens $w_1, w_2, \dots, w_n$ of the sentence to a forward \lstm, and (in reverse order) to a backward \lstm, obtaining the forward and backward hidden states ($\overrightarrow{h}_{1}, \dots, \overrightarrow{h}_{n} \in \R^{300}$ and $\overleftarrow{h}_{1}, \dots \overleftarrow{h}_{n} \in \R^{300}$). The concatenation of the last states ($h = [\overrightarrow{h}_{n}; \overleftarrow{h}_{1}]$) is fed to a multinomial Logistic Regression (\lr) layer, which produces a probability per class.

\begin{figure}[h]
  \centering
  \includegraphics[width=0.95\columnwidth]{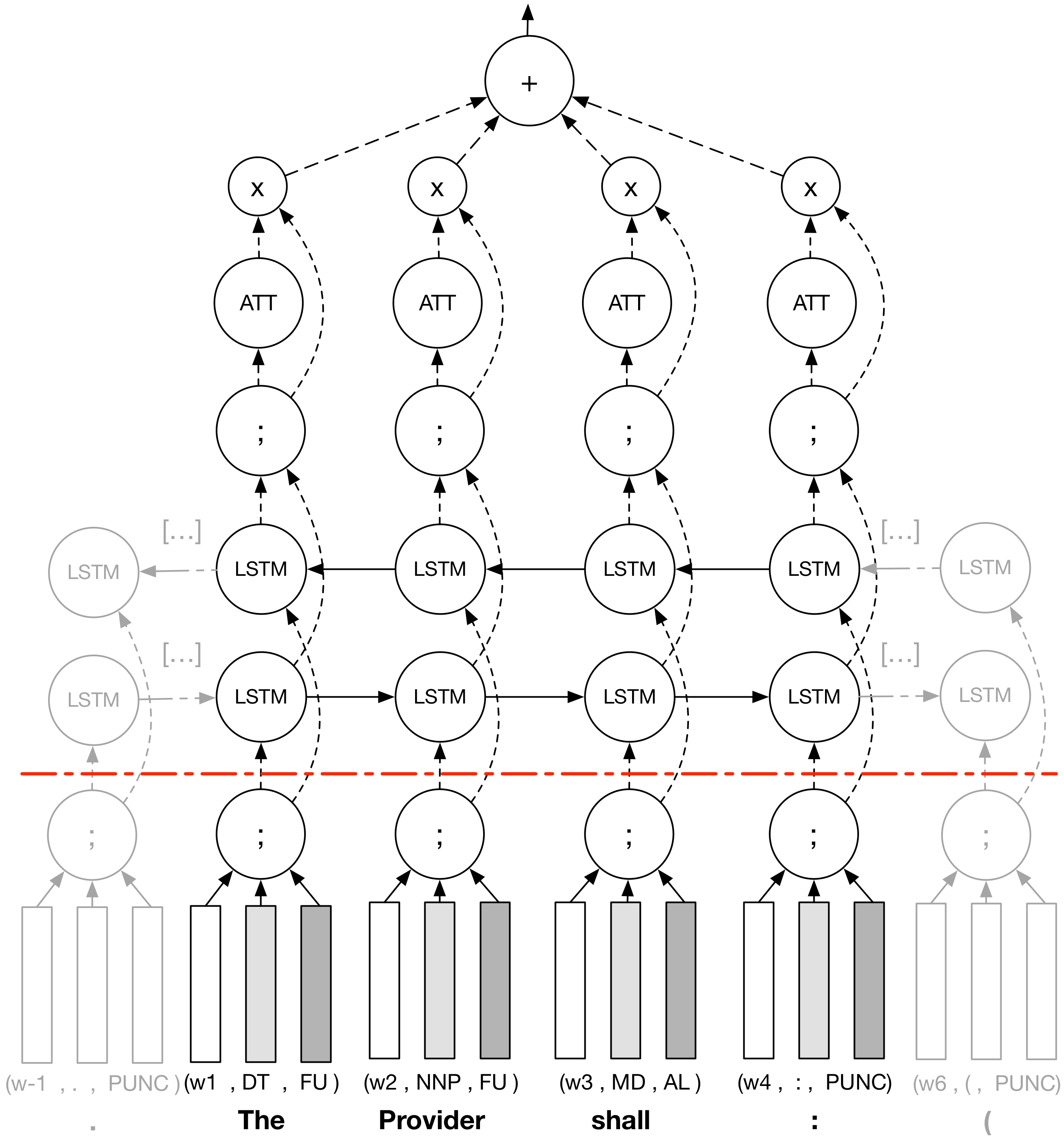}
  \caption{\bilstm with self-attention (\textsc{att} nodes) used on its own (\bilstmatt) or as the sentence encoder of the hierarchical \bilstm (\hbilstmatt, Fig.~\ref{fig:hierarchicalmodel}). In \xbilstmatt, the two \lstm chains also consider the words of surrounding sentences. The red dashed line is a drop-out layer.}
  \vspace*{-3mm}
  \label{fig:encoder}
\end{figure}
\label{sec:encoder}
\smallskip
\noindent\textbf{\bilstmatt}: When self-attention is added (Fig.~\ref{fig:encoder}), the sentence (or clause) is represented by the weighted sum ($h$) of the hidden states ($h_t = [\overrightarrow{h}_{t}; \overleftarrow{h}_{t}] \in \R^{600}$) of the \bilstm, where $a_1, \dots, a_n \in \R$ are attention scores, $v \in \R^{600}$, $b \in \R$:

\begin{eqnarray}
h &=& a_1 h_1 + \dots + a_t h_t + \dots + a_n h_n \label{eq:weightedSum}\\
a_t' &=& \tanh(v^T h_t + b) \\
a_t &=& \softmax(a_t'; a_1', \dots, a_n')
\end{eqnarray}
Again, $h$ is then fed to a multinomial \lr layer. Figure~\ref{fig:heatmapFlat} visualizes the attention scores ($a_1, \dots, a_n$) of \bilstmatt when reading some of the sentences (or clauses) of Table~\ref{tab:examples}. The attention scores are higher for modals, negations, words that indicate obligations or prohibitions (e.g., `obliged', `only'), and tokens indicating nested clauses (e.g., `(a)', `:', `;'), which allows \bilstmatt to focus more on these tokens (the corresponding states) when computing the sentence representation ($h$).

\smallskip
\noindent \textbf{\xbilstmatt}: In an extension of \bilstmatt, called \xbilstmatt, the \bilstm chain is fed with the token embeddings ($e_t$) not only of the sentence being classified, but also of the previous (and following) tokens (faded parts of Fig.~\ref{fig:encoder}), up to 150 previous (and 150 following) tokens, 150 being the maximum sentence length in the dataset.\footnote{Memory constraints did not allow including more tokens. We used a single \textsc{nvidia} 1080 \textsc{gpu}. All methods were implemented using \textsc{keras} (\url{https://keras.io/}) with a \textsc{tensorflow} backend (\url{https://www.tensorflow.org/}). We padded each sentence to the maximum length.}
This might allow the \bilstm chain to `remember' key parts of the surrounding sentences (e.g., a previous clause ending with `shall not:') when producing the context-aware embeddings (states $h_t$) of the current sentence. The self-attention mechanism still considers the states ($h_t$) of the tokens of the current sentence only, and the sentence representation ($h$) is still computed as in Eq.~\ref{eq:weightedSum}.

\begin{table*}[bht]
\centering
{\scriptsize
\begin{tabular}{|c|c|c|c|c|c|c|c|c|c|c|c|c|c|c|c|c|}
\hline
 & \multicolumn{4}{c|}{\textbf{\bilstm}} & \multicolumn{4}{c|}{\textbf{\bilstmatt}} & \multicolumn{4}{c|}{\textbf{\xbilstmatt}} & \multicolumn{4}{c|}{\textbf{\hbilstmatt}} \\ \hline
\textbf{Gold Class} 
 & P & R & F1 & AUC & P & R & F1 & AUC & P & R & F1 & AUC & P & R & F1 & AUC \\ \hline 
None & 0.95 & 0.91 & 0.93 & 0.98 & 0.97 & 0.90 & 0.93 & \cellcolor{mygray}\textbf{0.99} & 0.96 & 0.90 & 0.93  &0.98 & \cellcolor{mygray}\textbf{0.98} & \cellcolor{mygray}\textbf{0.96} & \cellcolor{mygray}\textbf{0.97} & \cellcolor{mygray}\textbf{0.99} \\ \hline
Obligation & 0.75 & 0.85 & 0.79 & 0.86  & 0.75 & 0.88 & 0.81 & 0.86 & 0.75 & 0.87 & 0.81 & 0.88  & \cellcolor{mygray}\textbf{0.87} & \cellcolor{mygray}\textbf{0.92} & \cellcolor{mygray}\textbf{0.90} & \cellcolor{mygray}\textbf{0.96} \\ \hline
Prohibition & 0.67 & 0.62 & 0.64 & 0.75  & 0.74 & 0.75 & 0.74 & 0.80 & 0.65 & 0.75 & 0.70 & 0.74  & \cellcolor{mygray}\textbf{0.84} & \cellcolor{mygray}\textbf{0.83} & \cellcolor{mygray}\textbf{0.84} & \cellcolor{mygray}\textbf{0.90} \\ \hline
Obl.\ List Begin & 0.70 & 0.86 & 0.77  & 0.81  & 0.71 & 0.85 & 0.77 & 0.83 & 0.72 & 0.75 & 0.74 & 0.80 & \cellcolor{mygray}\textbf{0.90} & \cellcolor{mygray}\textbf{0.89} & \cellcolor{mygray}\textbf{0.89} & \cellcolor{mygray}\textbf{0.93} \\ \hline
Obl.\ List Item & 0.53 & 0.66 & 0.59 & 0.64  & 0.48 & 0.70 & 0.57 & 0.60 & 0.49 & 0.78 & 0.60 & 0.66 & \cellcolor{mygray}\textbf{0.85} & \cellcolor{mygray}\textbf{0.94} & \cellcolor{mygray}\textbf{0.89} & \cellcolor{mygray}\textbf{0.94} \\ \hline
Proh.\ List Item & 0.59 & 0.35 & 0.43 & 0.50  & 0.61 & 0.55 & 0.59 & 0.62 & \cellcolor{mygray}\textbf{0.83} & 0.50 & 0.62 & 0.67  & 0.80 & \cellcolor{mygray}\textbf{0.84} & \cellcolor{mygray}\textbf{0.82} & \cellcolor{mygray}\textbf{0.92} \\ \hline
\textbf{Macro-average} & 0.70 & 0.70 & 0.70 & 0.74  & 0.73 & 0.78 & 0.74 & 0.78 & 0.73 & 0.76 & 0.73 & 0.79  & \cellcolor{mygray}\textbf{0.87} & \cellcolor{mygray}\textbf{0.90} & \cellcolor{mygray}\textbf{0.89} & \cellcolor{mygray}\textbf{0.94} \\ \hline
\textbf{Micro-average} & 0.90 & 0.88 & 0.88 & 0.94  & 0.90 & 0.88 & 0.89 & 0.96 & 0.90 & 0.88 & 0.89 & 0.94 & \cellcolor{mygray}\textbf{0.95} & \cellcolor{mygray}\textbf{0.95} & \cellcolor{mygray}\textbf{0.95} & \cellcolor{mygray}\textbf{0.98} \\ \hline
\end{tabular}
}
\caption{Precision, recall, \textsc{f1}, and \auc scores, with the best results in bold and gray background.}
\label{tab:evaluation}
\end{table*}

\begin{figure}[tb]
  \centering
  \includegraphics[width=0.95\columnwidth]{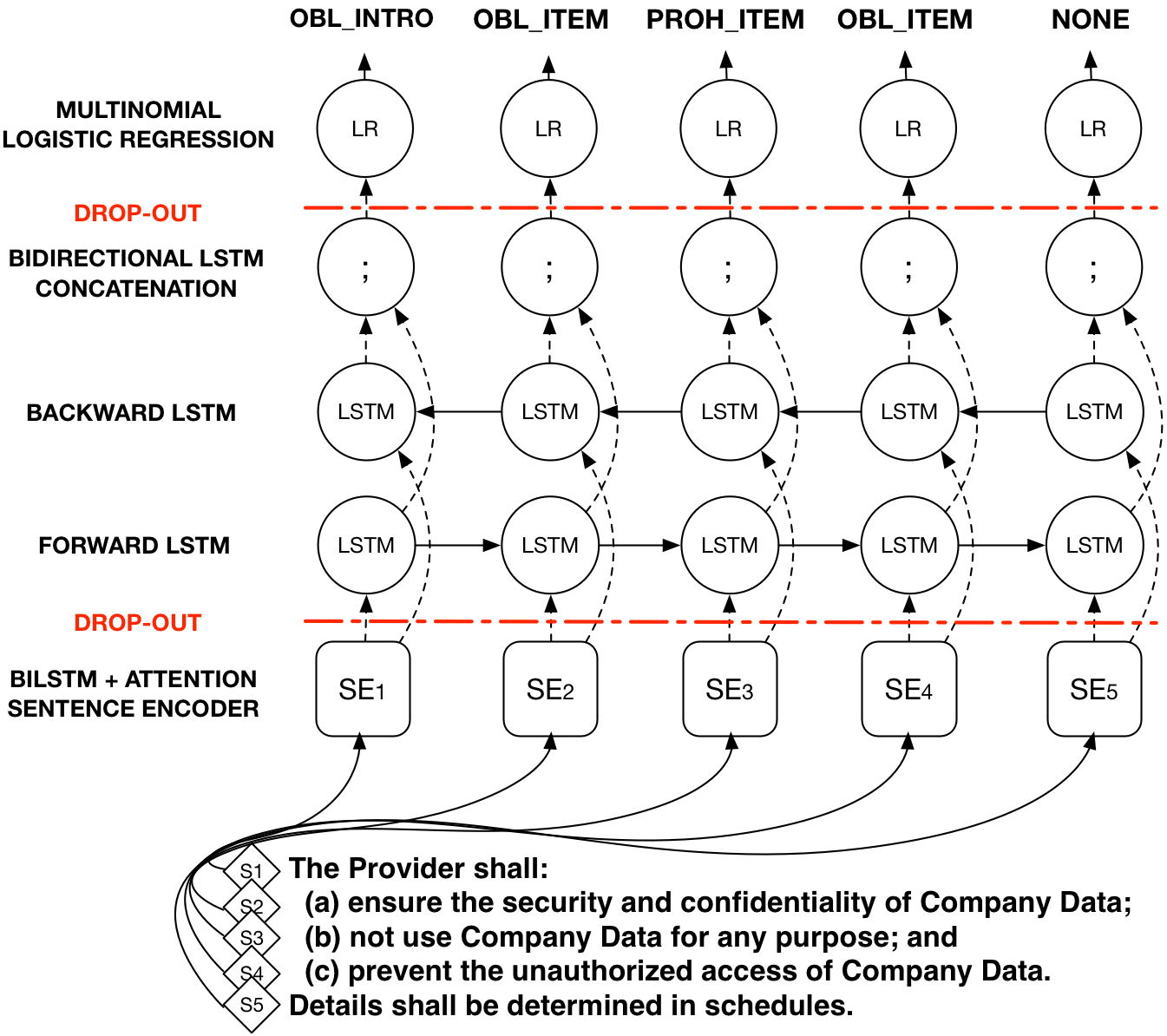}
  \caption{Upper part of the hierarchical \bilstm (\hbilstmatt). The sentence embeddings (\textsc{se}$_i$) are generated by the encoder of Fig.~\ref{fig:encoder}.}
  \label{fig:hierarchicalmodel}
\end{figure}

\smallskip
\noindent\textbf{\hbilstmatt}: The hierarchical \bilstm classifier, \hbilstmatt, considers all the sentences (or clauses) of an entire section. Each sentence (or clause) is first turned into a sentence embedding ($h \in \R^{600}$), as in \bilstmatt (Fig.~\ref{fig:encoder}). The sequence of sentence embeddings is then fed to a second \bilstm (Fig.~\ref{fig:hierarchicalmodel}), whose hidden states ($h^{(2)}_t = [\overrightarrow{h}^{(2)}_{t}; \overleftarrow{h}^{(2)}_{t}] \in \R^{600}$) are treated as context-aware sentence embeddings. The latter are passed on to a multinomial \lr layer, producing a probability per class, for each sentence (or clause) of the section. We hypothesized that \hbilstmatt would perform better, because it considers an entire section at a time, and salient information about a sentence or clause (e.g., that the opening clause of a list contains a negation or modal) can be `condensed' in its sentence embedding and interact with the sentence embeddings of distant sentences or clauses (e.g., a nested clause several clauses after the opening one) in the upper \bilstm (Fig.~\ref{fig:hierarchicalmodel}). 

\section{Experimental Results} \label{sec:experiments}

Hyper-parameters were tuned by grid-searching the following sets, and selecting the values with the best validation loss: \lstm hidden units \{100, 200, 300\}, batch size \{8, 16, 32\}, drop-out rate \{0.4, 0.5, 0.6\}. The red dashed lines of Fig.~\ref{fig:encoder}--\ref{fig:hierarchicalmodel} are drop-out layers.\footnote{We resample the drop-out mask at each time-step.}
We used categorical cross-entropy loss, Glorot initialization \cite{Glorot2010}, Adam \cite{Kingma2015}, learning rate 0.001, and early stopping on the validation loss. Table~\ref{tab:evaluation} reports the precision, recall, F1 score, area under the precision-recall curve (\auc) per class, as well as micro- and macro-averages. 

The self-attention mechanism (\bilstmatt) leads to clear overall improvements (in macro and micro \textsc{f1} and \auc, Table~\ref{tab:evaluation}) comparing to the plain \bilstm, supporting the hypothesis that self-attention allows the classifier to focus on indicative tokens. Allowing the \bilstm to consider tokens of neighboring sentences (\xbilstmatt) does not lead to any clear overall improvements. The hierarchical \hbilstmatt clearly outperforms the other three methods, supporting the hypothesis that considering entire sections and allowing the sentence embeddings to interact in the upper \bilstm (Fig.~\ref{fig:hierarchicalmodel}) is beneficial.

Notice that the three flat methods (\bilstm, \bilstmatt, \xbilstmatt) obtain particularly lower \textsc{f1} and \auc scores, compared to \hbilstmatt, in the classes that correspond to nested clauses (obligation list item, prohibition list item). This is due to the fact that the flat methods have no (or only limited, in the case of \xbilstmatt) view of the previous sentences, which often indicate if a nested clause is an obligation or prohibition (see, for example, examples 4--6 in Table~\ref{tab:examples}). 

\hbilstmatt is also much faster to train than \bilstm and \bilstmatt (Table~\ref{tab:time}), even though it has more parameters, because it converges faster (5-7 epochs vs.\ 12-15). \xbilstmatt is particularly slow, because its \bilstm processes the same sentences multiple times, when they are 
classified and when they are neighboring sentences.   

\begin{table}[h]
\centering
{\scriptsize
\begin{tabular}{|c|c|c|}
\hline
\textbf{Network} & 
\textbf{Training Time} & \textbf{Parameters} \\ \hline 
\textbf{\bilstm} &  5h 30m & 1,278M\\ \hline
\textbf{\bilstmatt} &  8h 30m & 1,279M\\ \hline
\textbf{\xbilstmatt} & 25h 40m & 1,279M\\ \hline
\textbf{\hbilstmatt} & 2h 30m & 1,837M\\ \hline
\end{tabular}
} 
\caption{Training times and parameters to learn.}
\vspace*{-3mm}
\label{tab:time}
\end{table}

\section{Related Work}

As already noted, we built upon the work of O'Neill et al.\ \shortcite{ONeill2017}. The dataset of O'Neill et al.\ contained financial legislation, not contracts, and was an order of magnitude smaller (obligations, prohibitions, permissions had 1,297 training, 622 test sentences in total, cf.\ Table~\ref{tab:dataset}), but also included permissions, which we did not consider.

Waltl et al.~\shortcite{Waltl2017} classified statements from German tenancy law into 22 classes
(including prohibition, permission, consequence), using active learning with Naive Bayes, \lr, \mlp classifiers, experimenting with 504 sentences.

Kiyavitskaya et al.\ \shortcite{Kiyavitskaya2008} used grammars, word lists, and heuristics to extract rights, obligations, exceptions, and other constraints from \textsc{us} and Italian regulations.

Asooja et al.\ \shortcite{Asooja2015} employed \textsc{svm}s with $n$-gram and manually crafted features to classify paragraphs of money laundering regulations into five classes (e.g., 
enforcement, monitoring, reporting), experimenting with 212 paragraphs.

In previous work \cite{Chalkidis2017,Chalkidis2017b} we focused on extracting contract elements (e.g., contractor names, legislation references, start and end dates, amounts), a task which is similar to named entity recognition. The best results were obtained by stacked \bilstm{s} \cite{Irsoy2014} or stacked \bilstm-\textsc{crf} models \cite{Ma2016}; hierarchical \bilstm{s} were not considered. By contrast, in this paper we considered obligation and prohibition extraction, treating it as a sentence (or clause) classification task, and showing the benefits of employing a hierarchical \bilstm model that considers both the sequence of words in each sentence and the sequence of sentences.

Yang et al.\ \shortcite{Yang2016} proposed a hierarchical \rnn with self-attention to classify texts. A first bidirectional \rnn turns the words of each sentence to a sentence embedding, and a second one turns the sentence embeddings to a document embedding, which is  fed to an \lr layer. Yang et al.\ use self-attention in both \rnn{s}, to assign attention scores to words and sentences. We classify sentences (or clauses), not entire texts, hence our second \bilstm does not produce a document embedding and does not use self-attention. Also, Yang et al.\ experimented with reviews and community question answering logs, whereas we considered legal texts. 

Hierarchical \rnn{s} have also been developed for multilingual text classification \cite{Pappas_IJCNLP_2017}, language modeling \cite{Lin2015}, and 
dialogue breakdown detection \cite{Xie2017}.

\section{Conclusions and Future Work}

We presented the legal text analytics task of detecting contractual obligations and prohibitions. We showed that self-attention improves the performance of a \bilstm classifier, the previous state of the art in this task, by allowing the \bilstm to focus on indicative tokens. We also introduced a hierarchical \bilstm (also using attention), which converts each sentence to an embedding, and then processes the sentence embeddings to classify each sentence. Apart from being faster to train, the hierarchical \bilstm outperforms the flat one, even when the latter considers the surrounding sentences, because the hierarchical model has a broader view of the discourse.

Further performance improvements may be possible by considering deeper self-attention mechanisms \cite{Pavlopoulos2017}, stacking \bilstm{s} \cite{Irsoy2014}, or pre-training the \bilstm{s} with auxiliary tasks \cite{Ramachandran2017}. The hierarchical \bilstm with attention of this paper may also be useful in other sentence, clause, or utterance classification tasks, for example in dialogue turn classification \cite{Xie2017}, detecting abusive user comments in on-line discussions \cite{Pavlopoulos2017}, and discourse segmentation \cite{Hearst1997}. We would also like to investigate replacing its \bilstm{s} with sequence-labeling \textsc{cnn}s \cite{Bai2018}, which may lead to efficiency improvements. 

\section*{Acknowledgments}
We are grateful to the members of \textsc{aueb}'s Natural Language Processing Group, for several suggestions that helped significantly improve this paper.

\bibliography{acl2018}

\begin{thebibliography}{23}
\expandafter\ifx\csname natexlab\endcsname\relax\def\natexlab#1{#1}\fi

\bibitem[{Alschnerd and Skougarevskiy(2017)}]{Alschnerd2017}
W.~Alschnerd and D.~Skougarevskiy. 2017.
\newblock {Towards an automated production of legal texts using recurrent
  neural networks}.
\newblock In \emph{Proceeding of the 16th International Conference on
  Artificial Intelligence and Law}, pages 159--168, London, UK.

\bibitem[{Ashley(2017)}]{Ashley2017}
K.D. Ashley. 2017.
\newblock \emph{Artificial Intelligence and Legal Analytics: New Tools for Law
  Practice in the Digital Age}.
\newblock Cambridge University Press.

\bibitem[{Asooja et~al.(2015)Asooja, Bordea, Vulcu, O'Brien, Espinoza,
  Abi-Lahoud, Buitelaar, and Butler}]{Asooja2015}
K.~Asooja, G.~Bordea, G.~Vulcu, L.~O'Brien, A.~Espinoza, E.~Abi-Lahoud,
  P.~Buitelaar, and T.~Butler. 2015.
\newblock Semantic annotation of finance regulatory text using multilabel
  classification.
\newblock In \emph{Proceedings of the International Workshop on Legal Domain
  and Semantic Web Applications}, Portoroz, Slovenia.

\bibitem[{Bai et~al.(2018)Bai, Kolter, and Koltun}]{Bai2018}
S.~Bai, J.Z. Kolter, and V.~Koltun. 2018.
\newblock An empirical evaluation of generic convolutional and recurrent
  networks for sequence modeling.
\newblock \emph{CoRR}, abs/1803.01271.

\bibitem[{Chalkidis and Androutsopoulos(2017)}]{Chalkidis2017b}
I.~Chalkidis and I.~Androutsopoulos. 2017.
\newblock A deep learning approach to contract element extraction.
\newblock In \emph{Proceedings\ of the 30th International Conference\ on Legal
  Knowledge and Information Systems}, pages 155--164, Luxembourg.

\bibitem[{Chalkidis et~al.(2017)Chalkidis, Androutsopoulos, and
  Michos}]{Chalkidis2017}
I.~Chalkidis, I.~Androutsopoulos, and A.~Michos. 2017.
\newblock Extracting contract elements.
\newblock In \emph{Proceedings\ of the 16th International Conference\ on
  Artificial Intelligence and Law}, pages 19--28, London, UK.

\bibitem[{Glorot and Bengio(2010)}]{Glorot2010}
X.~Glorot and Y.~Bengio. 2010.
\newblock Understanding the difficulty of training deep feedforward neural
  networks.
\newblock In \emph{Proceedings of the International Conference on Artificial
  Intelligence and Statistics}, pages 249--256, Sardinia, Italy.

\bibitem[{Graves et~al.(2013)Graves, Jaitly, and Mohamed}]{Graves2013}
A.~Graves, N.~Jaitly, and A.~Mohamed. 2013.
\newblock Hybrid speech recognition with deep bidirectional {LSTM}.
\newblock In \emph{IEEE Workshop on Automatic Speech Recognition and
  Understanding}, pages 273--278, Olomouc, Czech Republic.

\bibitem[{Hearst(1997)}]{Hearst1997}
Marti~A. Hearst. 1997.
\newblock Texttiling: Segmenting text into multi-paragraph subtopic passages.
\newblock \emph{Computational Linguistics}, 23(1):33--64.

\bibitem[{Irsoy and Cardie(2014)}]{Irsoy2014}
O.~Irsoy and C.~Cardie. 2014.
\newblock Deep recursive neural networks for compositionality in language.
\newblock In \emph{Proceedings of the 27th International Conference on Neural
  Information Processing Systems}, pages 2096--2104, Montreal, Canada.

\bibitem[{Kim and Goebel(2017)}]{Kim2017}
M.Y. Kim and R.~Goebel. 2017.
\newblock Two-step cascaded textual entailment for legal bar exam question
  answering.
\newblock In \emph{Proceedings of the 4th Competition on Legal Information
  Extraction/Entailment}, London, UK.

\bibitem[{Kingma and Ba(2015)}]{Kingma2015}
D.~P. Kingma and J.~Ba. 2015.
\newblock Adam: A method for stochastic optimization.
\newblock In \emph{Proceedings of the 5th International Conference on Learning
  Representations}, San Diego, CA, USA.

\bibitem[{Kiyavitskaya et~al.(2008)Kiyavitskaya, Zeni, Breaux, Ant{\'o}n,
  Cordy, Mich, and Mylopoulos}]{Kiyavitskaya2008}
N.~Kiyavitskaya, N.~Zeni, Travis~D. Breaux, Annie~I. Ant{\'o}n, James~R. Cordy,
  L.~Mich, and J.~Mylopoulos. 2008.
\newblock Automating the extraction of rights and obligations for regulatory
  compliance.
\newblock In \emph{Proceedings of the 27th International Conference on
  Conceptual Modeling}, pages 154--168, Barcelona, Spain.

\bibitem[{Lin et~al.(2015)Lin, Liu, Yang, Li, Zhou, and Li}]{Lin2015}
R.~Lin, S.~Liu, M.~Yang, M.~Li, M.~Zhou, and S.~Li. 2015.
\newblock Hierarchical recurrent neural network for document modeling.
\newblock In \emph{Proceedings of the Conference on Empirical Methods in
  Natural Language Processing}, pages 899--907, Lisbon, Portugal.

\bibitem[{Ma and Hovy(2016)}]{Ma2016}
X.~Ma and E.~Hovy. 2016.
\newblock End-to-end sequence labeling via bi-directional {LSTM}-{CNN}s-{CRF}.
\newblock In \emph{Proceedings of the 54th Annual Meeting of the ACL}, pages
  1064–--1074, Berlin, Germany.

\bibitem[{Mikolov et~al.(2013)Mikolov, Sutskever, Chen, Corrado, and
  Dean}]{MikolovSCCD13}
T.~Mikolov, I.~Sutskever, K.~Chen, G.~Corrado, and J.~Dean. 2013.
\newblock Distributed {R}epresentations of {W}ords and {P}hrases and their
  {C}ompositionality.
\newblock In \emph{Proceedings of the 26th International Conference on Neural
  Information Processing Systems}, Stateline, NV.

\bibitem[{O'Neill et~al.(2017)O'Neill, Buitelaar, Robin, and
  Brien}]{ONeill2017}
J.~O'Neill, P.~Buitelaar, C.~Robin, and L.~O' Brien. 2017.
\newblock {Classifying Sentential Modality in Legal Language: A Use Case in
  Financial Regulations, Acts and Directives}.
\newblock In \emph{Proceedings of the 16th International Conference on
  Artificial Intelligence and Law}, pages 159--168, London, UK.

\bibitem[{Pappas and Popescu-Belis(2017)}]{Pappas_IJCNLP_2017}
N.~Pappas and A.~Popescu-Belis. 2017.
\newblock Multilingual hierarchical attention networks for document
  classification.
\newblock In \emph{Proceedings of the 8th International Joint Conference on
  Natural Language Processing}, Tapei, Taiwan.

\bibitem[{Pavlopoulos et~al.(2017)Pavlopoulos, Malakasiotis, and
  Androutsopoulos}]{Pavlopoulos2017}
J.~Pavlopoulos, P.~Malakasiotis, and I.~Androutsopoulos. 2017.
\newblock Deeper attention to abusive user content moderation.
\newblock In \emph{Proceedings of the Conference on Empirical Methods in
  Natural Language Processing}, pages 1125--1135, Copenhagen, Denmark.

\bibitem[{Ramachandran et~al.(2017)Ramachandran, Liu, and
  Le}]{Ramachandran2017}
P.~Ramachandran, P.r~J. Liu, and Q.~V. Le. 2017.
\newblock Unsupervised pretraining for sequence to sequence learning.
\newblock In \emph{Proceedings of the Conference on Empirical Methods in
  Natural Language Processing}, pages 383--391, Copenhagen, Denmark.

\bibitem[{Waltl et~al.(2017)Waltl, Muhr, Glaser, Bonczek, Scepankova, and
  Matthes}]{Waltl2017}
B.~Waltl, J.~Muhr, I.~Glaser, G.~Bonczek, E.~Scepankova, and F.~Matthes. 2017.
\newblock Classifying legal norms with active machine learning.
\newblock In \emph{Proceedings\ of the 30th International Conference\ on Legal
  Knowledge and Information Systems}, pages 11--20, Luxembourg City‚
  Luxembourg.

\bibitem[{Xie and Ling(2017)}]{Xie2017}
Z.~Xie and G.~Ling. 2017.
\newblock Dialogue breakdown detection using hierarchical bi-directional
  {LSTMs}.
\newblock In \emph{Proceedings of the 6th Dialog System Technology Challenges
  (Track 3: Dialog Breakdown Detection)}, Long Beach, USA.

\bibitem[{Yang et~al.(2016)Yang, Yang, Dyer, He, Smola, and Hovy}]{Yang2016}
Z.~Yang, D.~Yang, C.~Dyer, X.~He, A.~Smola, and E.~Hovy. 2016.
\newblock Hierarchical attention networks for document classification.
\newblock In \emph{Proceedings of the 15th Conference of the North American
  Chapter of the Association for Computational Linguistics: Human Language
  Technologies}, pages 1480--1489, San Diego, CA, USA.

\end{thebibliography}
\bibliographystyle{acl_natbib}

\end{document}